\documentclass{article}
\usepackage{spconf,amsmath,graphicx}
\usepackage{stfloats}

\usepackage[ruled,linesnumbered]{algorithm2e}

\usepackage{booktabs} 
\usepackage{multirow} 

\usepackage{enumitem}
\setenumerate[1]{itemsep=0pt,partopsep=0pt,parsep=\parskip,topsep=3pt}
\setitemize[1]{itemsep=0pt,partopsep=0pt,parsep=\parskip,topsep=3pt}
\setdescription{itemsep=0pt,partopsep=0pt,parsep=\parskip,topsep=3pt}

\usepackage{cite}

\usepackage{hyperref}
\hypersetup{hypertex=true,
            colorlinks=true,
            linkcolor=blue,
            anchorcolor=blue,
            citecolor=blue}

\title{ADAPTIVE DATA AUGMENTATION FOR CONTRASTIVE LEARNING}
\name{Yuhan Zhang$^{1,2}$, He Zhu$^{1,3}$, Shan Yu$^{1,2,3}$\sthanks{Corresponding author. (e-mail: shan.yu@nlpr.ia.ac.cn)}}
\address{$^1$ Brainnetome Center, National Laboratory of Pattern Recognition (NLPR), \\
Institute of Automation, Chinese Academy of Sciences (CASIA), Beijing, China\\
$^2$ School of Artificial Intelligence, University of Chinese Academy of Sciences (UCAS), Beijing, China\\
$^3$ School of Future Technology, University of Chinese Academy of Sciences (UCAS), Beijing, China
}

\begin{document}
%
\maketitle
\begin{abstract}
In computer vision, contrastive learning is the most advanced unsupervised learning framework. Yet most previous methods simply apply fixed composition of data augmentations to improve data efficiency, which ignores the changes in their optimal settings over training. Thus, the pre-determined parameters of augmentation operations cannot always fit well with an evolving network during the whole training period, which degrades the quality of the learned representations. In this work, we propose AdDA, which implements a closed-loop feedback structure to a generic contrastive learning network. AdDA works by allowing the network to adaptively adjust the augmentation compositions according to the real-time feedback. This online adjustment helps maintain the dynamic optimal composition and enables the network to acquire more generalizable representations with minimal computational overhead. AdDA achieves competitive results under the common linear protocol on ImageNet-100 classification (+1.11$\%$ on MoCo v2).
\end{abstract}
\section{Introduction}
\label{sec:intro}

Several recent studies show that deep networks can achieve similar or even better results than supervised training on many downstream tasks\cite{res1,res2,res3,res4,res5,res6}. In computer vision, contrastive learning proves to be a highly effective unsupervised learning framework. Previous work has reported that data augmentation is the key for self-supervised training\cite{res7,res8,res9}. The SimCLR approach of Chen et al.\cite{res1} shows that data augmentation composition performs better than individual operators, which was re-confirmed later with modifications made on the MoCo framework\cite{res10} (named MoCo v2 \cite{res11}). Peng et al. proposed ContrastiveCrop\cite{res12}, which helps to generate better crops for Simese representation learning.

\begin{figure}[htb]
  \centering
  \setlength{\abovecaptionskip}{-0.2 cm}
  \includegraphics{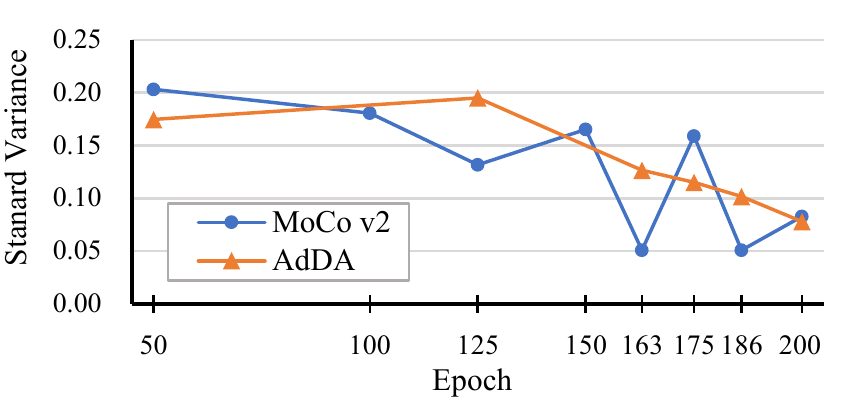}
\caption{The standard variance of different data augmentations for the same encoder on ImageNet-100. We present the standard variance of the classification accuracy for the four augmentations used in MoCo v2’s pretext task. Technical details are shown in section \ref{sec:3}.}
\label{fig1}
\end{figure}

Despite different techniques, most recent methods conduct pretext task on a particular data augmentation composition, and rarely concerned about adjusting the augmentations on-the-fly during training. Those pre-determined augmentation policies may reduce the training efficiency, as the optimal parameters of data augmentations are continually changing with time. For example, the fixed augmentations may fit well with the network in the beginning, but perform poorly when it comes to the next training period. In other words, augmentation operations with those fixed settings may be sub-optimal to generate good representations.

We test four operators in Fig. \ref{fig1} to illustrate the issue of dynamic optimal parameters for data augmentations. Notably, the figure shows the existence of variance in their increasing fitness with the network when no feedback is implemented. The fluctuation suggests that training with a particular composition may limit the performance of the network and thus highlights the necessity of online adjustment.

Given this, here we propose AdDA method to introduce a strategy to help the network find the optimal composition for the current situation, thereby improving the generalization performance of the network. AdDA works by sampling sub-batches with negligible training overhead, and improves MoCo v2 by 1.11$\%$ classification accuracy on Imagenet-100. Our main contributions are summarized as follows: 
\begin{figure*}[htbp]
\centering
\setlength{\abovecaptionskip}{-0.2 cm}
\includegraphics[width=14.7cm]{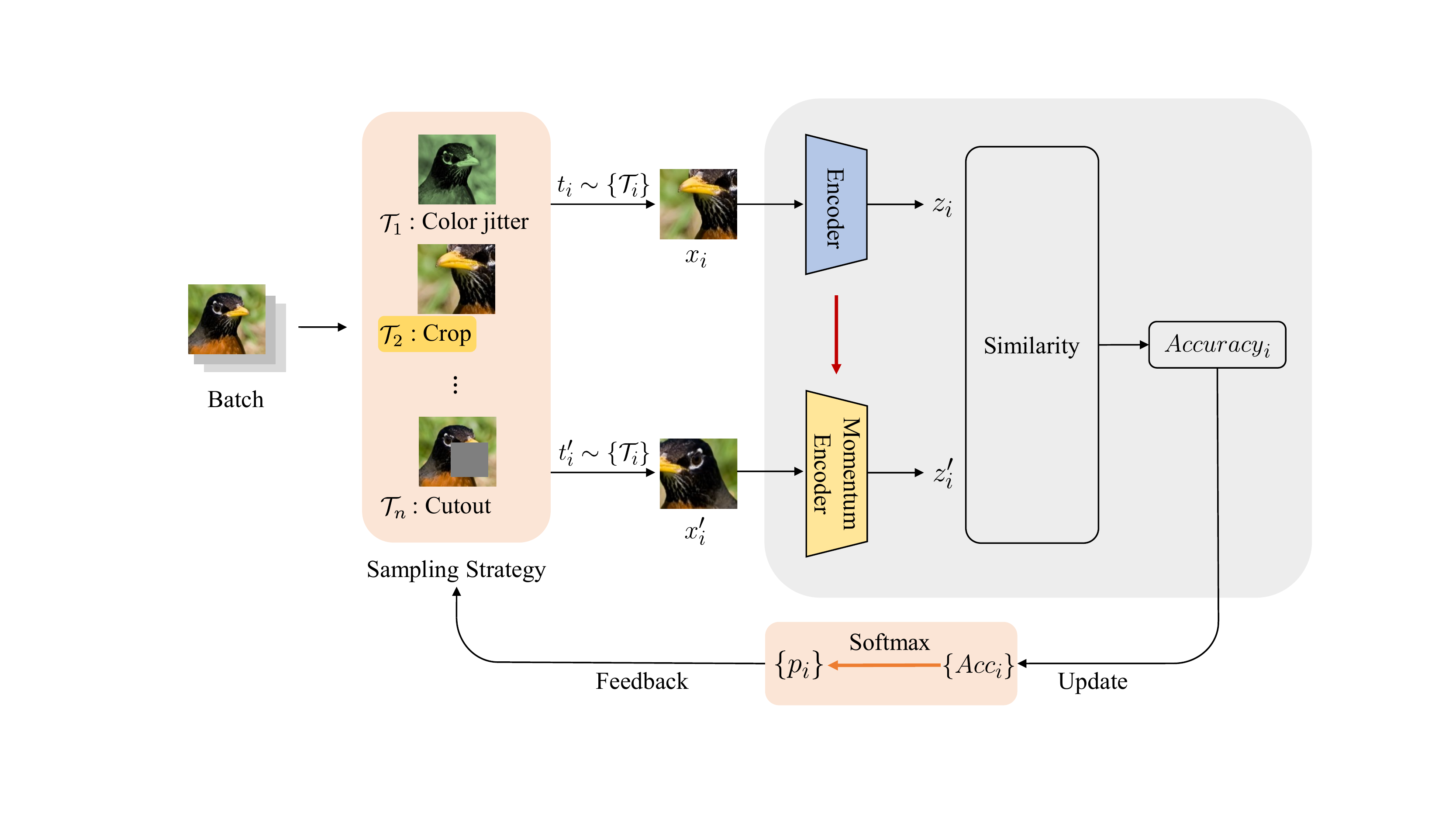}
\caption{The pipeline of AdDA. $\mathcal{T}_i$ is the random data augmentation operator sampled from ${\{\mathcal{T}_i\}}_{i=1}^N$ with the sampling probability of $p_i$. All the augmented data examples share the same encoder and momentum encoder. Similarities of the representation vectors are measured to compute $\mathcal{T}_i$’s average pretext task accuracy $Acc_i$ and ${\{Acc_i\}}_{i=1}^N$ are used to update ${\{p_i\}}_{i=1}^N$, which are sent back to the \textit{Sampling Strategy} module for the next epoch training.} 
\label{pipeline}
\end{figure*}

\begin{itemize}
\item We focus on the dynamic changes in optimal settings for different augmentations, and propose AdDA that helps the network acquire more generalizable representations with online adjustments.
\item As a basic sampling method, AdDA enables the network to adaptively adjust the data augmentation compositions according to the real-time feedback, which avoids training with globally fixed parameters and thus improves data efficiency.
\item Without increasing the time complexity, AdDA is useful when searching for the dynamically changing parameters during training. In addition, we show that the optimal distribution of those parameters varies among different data augmentations.
\end{itemize}

\section{RELATION TO PRIOR WORK}
\label{sec:format}

As a milestone in contrastive learning, SimCLR\cite{res1} shows that stronger data augmentations help to bring accuracy gains with contrastive learning and introduced a nonlinear transformation. MoCo v2\cite{res11}, which we use to implement our algorithm, adds stronger data augmentations and the same nonlinear projection head as SimCLR.

One of the key issues in contrastive learning is to design positives selection\cite{res12}. In this case, strong data augmentations\cite{res1,res13,res14} adapted from works in supervised training\cite{res15,res16,res17,res18,res19,res20} are often applied to generate positive pairs. Yet less attention has been paid to the dynamic best choice for data augmentations, which we find may reduce the training efficiency of the network. 

Instead of finding the global data augmentation policy\cite{res15}, AdDA enables the network to adaptively learn from images transformed by different compositions. In addition, our method helps the network to acquire better representations with online adjustment, considering that the optimal parameters of different augmentations various and changes over training.

\section{METHODOLOGY}
\label{sec:3}

Data augmentation compositions are used in our method, as single transformation can hardly suffice to learn good representations\cite{res1}. We formulate the problem of adjusting augmentation compositions as a discrete search problem and update their sampling weight every epoch according to the accuracy feedback of the pretext task. As illustrated in Fig. \ref{pipeline}, the training phase of AdDA comprises the following three parts:

\noindent\textbf{Sampling strategy.} We randomly sample \textit{N} sub-batches with probabilities of  $p_1,p_2,…,p_N$, each of which is transformed by a stochastic data augmentation module where $\mathcal{T}_i$ is applied to the \textit{i}’th sub-batch. AdDA focuses on providing the network with alternative compositions rather than a complete search space, and thus we use \textit{N} of 3, 6 and 7 to show the effect of online adjustment. Other values of \textit{N} may have better results.

\begin{algorithm}
\caption{AdDA ’s main learning algorithm}\label{algorithm}
\SetKwData{Left}{left}\SetKwData{This}{this}\SetKwData{Up}{up}
  \SetKwFunction{Union}{Union}\SetKwFunction{FindCompress}{FindCompress}
  \SetKwInOut{Input}{input}\SetKwInOut{Output}{output}

  \Input{training samples $X$, constant $\tau$, parameter sets $Args$, updating rate $ur$}
  \Output{encoder networks}
initialize sampling probabilities ${\{p_i\}}_{i=1}^N$\:
\vspace{-10pt}
\begin{equation}
    p_1=p_2=…=p_N
    \nonumber
\end{equation}\\
\vspace{-10pt}
\For{\rm{\textbf{all}} \rm{training epoch}}{
compute the size of each sub-batch:
\vspace{-10pt}
\begin{equation}
    {num\_data}_i=Softmax(p_i\times ur)\times num\_X
    \nonumber
\end{equation}\\
\vspace{-10pt}
update samplers and resample sub-batches\;

\For{\rm{\textbf{all}} \rm{sub-batches}}{
draw two augmentation functions $ \mathcal{T}_i$ and $\mathcal{T}'_i$\;
transform and map the training example\;
compute $\mathcal{L}_z$ and measure similarity\;
update networks to minimize $\mathcal{L}_z$\;
save the pretext task accuracy ${acc}_i$;
}
update sampling probability for each sub-batch:
\vspace{-10pt}
\begin{equation}
    p_i^{(t+1)}=mean(1-{acc}_i^t)
    \nonumber
\end{equation}\\
\vspace{-10pt}
}
\end{algorithm}

\noindent\textbf{Contrastive learning.} Random cropping is applied to take two random views of the same image (denoted as $x_i$ and $x'_i$), which are encoded by their own encoders respectively. The extracted representation vectors $z_i$ and $z'_i$ are used to measure the similarity, where dot product is computed to provide the accuracy of pretext task. We then consider infoNCE\cite{res3} to compute the loss of each composition:

\vspace{-15pt}
\begin{equation}
     \ell_{z_i,z'_i,Q}=-{\rm log}\frac{{\rm exp}(z_i\cdot z'_i/\tau)}{{\rm exp}(z_i\cdot z'_i/\tau)+\sum_{k\in Q}{\rm exp}(z_i\cdot k/\tau)}  
     \tag{1} 
\end{equation}
\vspace{-10pt}

\noindent where $\tau$ is a temperature parameter and all the embeddings are $\ell_2$ normalized. The final loss $\mathcal{L}_z$, which is computed across all compositions, is given by:

\vspace{-9pt}
\begin{equation}
    \mathcal{L}_z=
    \begin{matrix} 
    \sum_{i=1}^{n} \ell_{z_i,z'_i,Q}p_i
    \end{matrix} 
    \tag{2} 
\end{equation}
\vspace{-15pt}

\noindent Here the loss $\mathcal{L}_z$ allows the encoder networks to keep track of the data examples in all sub-batches, as they share the same key encoder and query encoder.

\noindent\textbf{Feedback.} The core of AdDA is to make use of the accuracy feedback. In the first epoch, we initialize the sampling probabilities to ensure a fair assignment. Namely, we always assume that all the compositions are of the same significance to the network and thus the same number of images are assigned to each sub-batch in the beginning. A softmax function is then used to update the sampling probabilities in the $t+1$ epoch: 

\vspace{-10pt}
\begin{equation}
    \label{qua3}
    p^{t+1}=Softmax((1-Acc^t )\times ur)\tag{3} 
\end{equation}
\vspace{-15pt}

\noindent where \textit{ur} is the updating rate. We hypothesize that it is easier to extract useful features from compositions with better pretext task accuracy and, in order to improve the quality of representations, decrease the size of those sub-batches in the next epoch according to the updating rate. Algorithm \ref{algorithm} summarizes the training phase of AdDA.

\begin{table}[h!]
\vspace{-10pt}
  \begin{center}

    \caption{Comparison under linear classification protocol on ImageNet-100. We first focus on the applied frequency of color jittering and bold the highest result as well as the final frequency. Here the final frequency denotes the composition that has the largest sub-batch in the last epoch. Results that present standard deviation were averaged over at least 2 runs.}
    \label{table1} 
    \vspace{2pt}
    \begin{tabular}{l|l|l} 
      \toprule  
      {Method} & {$f_{\rm Jitter}$} & {Top1 $\% (\pm\sigma)$}\\
      \midrule
      MoCo & 0.8 & 74.7\\
      MoCo v2 & 0.8 & 77.5\\
        & 0.6 & 77.72\\
        & 0.7 & 77.78\\
        \midrule
        
        \multicolumn{3}{l} {\textit{Using 3 data augmentation compositions:}}\\
        \midrule
        MoCo v2 &$\qquad$ \\
        $\hspace{0.7cm}$ + AdDA & (\textbf{0.6}, 0.8, 1)&77.82 ($\pm$ 0.02)\\
        & (\textbf{0.6}, 0.7, 0.8)& \textbf{77.99} ($\pm$ 0.07)\\
        \midrule
        \multicolumn{3}{l} {\textit{Using 6 data augmentation compositions:}}\\
        \midrule
         MoCo v2 & $\qquad$\\
        \hspace{0.7cm} + AdDA & (0, 0.2, 0.4, + & {\multirow{2}{*}{77.53 ($\pm$0.09)}}\\
        & 0.6, \textbf{0.8}, 1)\\
      \bottomrule 
    \end{tabular}
  \end{center}
  \vspace{-10pt}
  
\end{table}

\section{EXPERIMENT}
\label{sec:typestyle}
\subsection{Technical details}
\label{Technical details}
\vspace{-8pt}
One of the important parameters for augmentations is the frequency that they are applied with, which we use in our method to investigate the effect of dynamic adjustment. We conduct several experiments by adjusting the applied frequency of random color jittering, random grayscale conversion, random gaussian blur, and random horizontal flip. All the augmentations are available in PyTorch’s torchvision package.

\noindent\textbf{ImageNet-100.} We train AdDA on a subset of the common ImageNet-1k (ImageNet-100), which is also used in \cite{res21} to discuss variance. For ImageNet-100, the training set consists of 100 classes, each of which contains an average of 1000 images, \textit{i.e.}, 126,689 images in total, while each class in the validation set contains 50 images.

\noindent\textbf{Default setting.} For ImageNet-100, we use a batch size of 128, an initial learning rate of 0.03, and a temperature parameter of 0.2. The pretext task trains for 200 epochs on 4 GPU servers. The default frequencies of the four augmentations are 0.8, 0.2, 0.5 and 0.5 respectively, while the default updating rate is 1.0.
\vspace{-15pt}
\subsection{Linear Classification Protocol}
\label{Linear Classification Protocol}
\vspace{-8pt}
Following the common protocol, we verify our method by linear classification on frozen features of a ResNet\cite{res22} , with an initial learning rate of 30 and a batch size of 256. The linear classifier trains for 100 epochs using 2 GPU servers. Experiments of ablation (shown in Fig. \ref{fig1}) use the same classifier except for different augmentations. As shown in Table \ref{table1}, we report top-1 classification accuracy on ImageNet-100 validation set.

\begin{table}[h!]
\vspace{-10pt}
  \begin{center}
    \caption{Comparison under linear classification protocol on ImageNet-100. Results with standard deviation were measured by 3 runs.}
    \label{table2}
    \vspace{2pt}
    \begin{tabular}{l|l|l} 
      \toprule  
      {Method} & {\textit{ur}} & {Top1 $\% (\pm\sigma)$}\\
      \midrule
      MoCo & 1 & 74.7\\
      MoCo v2 & 1 & 77.5\\
       \midrule
        MoCo v2 + AdDA & 1 &78.32 ($\pm$ 0.27)\\
        & 0.8 & \textbf{78.61} ($\pm$ 0.48)\\
        & 1.2 & 78.27 ($\pm$ 0.13)\\
      \bottomrule 
    \end{tabular}
  \end{center}
  \vspace{-10pt}
 
\end{table}

\noindent\textbf{Active learning mechanism.} We first perform unsupervised pre-training by tuning the applied frequency of color jittering, while parameters of the others remain unchanged. In experiments, we center on MoCo’s setting (0.8) to set our frequencies (0.6, 0.8, 1.0) and show accuracy gains over MoCo v2: results are on average about 0.32$\%$ higher.

Notably, the optimal applied frequency in the last epoch is 0.8 when using 6 compositions, proving that MoCo’s settings are quite reliable. However, compared with those using 3 compositions, the performance of experiments with 6 compositions decreases by 0.23$\%$. In addition, we observe that training with frequencies of 0.6, 0.7 and 0.8 seems to perform consistently better than that of other frequencies (e.g., 0.6, 0.8 and 1.0), which suggests that there exists an optimal search space for augmentation compositions. 

Experiments show that the final frequency of color jittering varies from 0.6 to 0.8. Intuitively, the pretext task becomes harder with the increase of frequency, which tends to provide stronger compositions. However, higher frequency does not always bring accuracy benefits, as overly complex task may limit the generality of the network. This helps to explain the reason why $f_{\rm Jitter}=1$ fails to become the final composition in all our experiments. Thus, although strong data augmentations are needed to obtain effective features, simply increasing the applied frequency may hurt the generalization performance of the network. 

Individual augmentation compositions are used to verify the effects of our method. As shown in Table \ref{table1}, we observe that no single composition is able to surpass the performance of AdDA, even though we set the applied frequency within the optimal range. In experiments, the input images are transformed by two individual compositions respectively, with $f_{\rm Jitter}$ of 0.6 and 0.7. We see that the network gives consistent accuracy gains over MoCo v2 (+0.22$\%$, +0.28$\%$), but fails to outperform AdDA configurations.

We then perform several experiments with the remaining three augmentations using default updating rate. We find that those augmentations tend to have larger optimal search space with improved accuracy. The averaged results all surpass MoCo v2 with the highest being 78.32$\%$ (+0.82$\%$).

\begin{figure}[htb]
  \centering
  \includegraphics{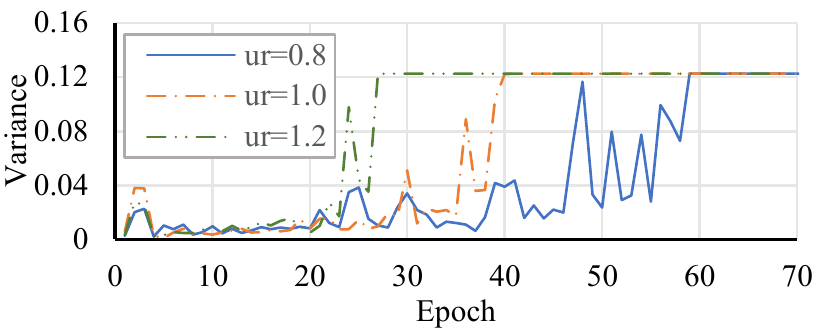}
  \vspace{-10pt}
\caption{Learning with different updating rate. Here we show the variance of sampling probability for seven data augmentation compositions in the first 70 epochs.}
\label{fig2}
\end{figure}

\noindent\textbf{Further study on updating rate.} We observe that the net work tends to learn better representations when training within the optimal range and thus the searching period matters.

Here we introduce \textit{ur} (shown in Eqn. \ref{qua3}) as a hyperparameter to adjust the updating rate of sampling weight for each composition, and thus speed up or slow down the process of dynamic adjustment. To further demonstrate the importance of ur, we therefore study the performance of AdDA by adjusting the applied frequency of random gaussian blur with 7 augmentation compositions, which shows to be the best pretraining setting in our previous experiments. 

Fig. \ref{fig2} visualizes the training processes with different updating rate. Notably, the variance of sampling probability shows an upward trend and fluctuates just before reaching the final stage. We conclude that the network tends to better “understand” the difference between various augmentation compositions, and thus widens their gap during training. The fluctuation suggests that AdDA helps to prevent the network from obtaining premature parameters.We report the effect of tuning updating rate in Table \ref{table2}.

\section{CONCLUSION}
\label{sec:majhead}

In this paper, we propose AdDA, which is tailored to provide adaptive data augmentation compositions for the network. AdDA implements a closed-loop feedback structure to a generic contrastive learning framework, and adjust the sampling probabilities on-the-fly according to the accuracy feedback of pretext task. We focus on the dynamic changes in parameters, and show the effect of training with optional compositions. Our method has shown positive results on classification task and improves the generalization performance of the network. We hope that AdDA could inspire further exploration of online adjustments.



\section{ACKNOWLEDGMENT}
\label{sec:print}

This work was supported in part by the STI 2030—Major Project (2021ZD0200402), the International Partnership Program of the Chinese Academy of Sciences (CAS) (173211KYSB20200021), and the Strategic Priority Research Program of CAS (XDB32040200).

\vfill\pagebreak



\bibliographystyle{IEEEbib}

\bibliography{ref.bib}
\end{document}